\documentclass{mva_style}
\usepackage{amsmath, amssymb, amsfonts}  
\usepackage{graphicx}                    
\usepackage{booktabs}                    
\usepackage{xcolor}                      
\usepackage{algorithm}                   
\usepackage{algpseudocode}               
\usepackage{multirow}                    
\usepackage{hyperref}                    
\usepackage{authblk}                     
\usepackage{wrapfig}                     
\usepackage{subcaption}                  
\usepackage{cite}                        
\usepackage{adjustbox}                   
\usepackage{bbding}                      
\usepackage{pifont}                      
\usepackage{enumitem}

\usepackage{float}

\usepackage[sort&compress,numbers]{natbib}
\usepackage{etoolbox}
\usepackage[font=scriptsize,labelfont=scriptsize]{caption}

\finalcopy 

\begin{document}
\title{Parallel Sampling of Diffusion Models on $SO(3)$}


\author{Yan-Ting Chen, Hao-Wei Chen, Tsu-Ching Hsiao, and Chun-Yi Lee}
\affil{Elsa Lab, Department of Computer Science and Information Engineering \\ National Taiwan University, Taiwan}

\maketitle
\pagestyle{plain}

\section*{\centering Abstract}
\textit{
In this paper, we design an algorithm to accelerate the diffusion process on the $SO(3)$ manifold. The inherently sequential nature of diffusion models necessitates substantial time for denoising perturbed data. To overcome this limitation, we proposed to adapt the numerical Picard iteration for the $SO(3)$ space. We demonstrate our algorithm on an existing method that employs diffusion models to address the pose ambiguity problem. Moreover, we show that this acceleration advantage occurs without any measurable degradation in task reward. The experiments reveal that our algorithm achieves a speed-up of up to $4.9\times$, significantly reducing the latency for generating a single sample.
}
\vspace{-0.7em}
\section{Introduction}
\vspace{-0.7em}

Diffusion generative models \cite{ho2020denoisingdiffusionprobabilisticmodels, rombach2022highresolutionimagesynthesislatent, hsiao2024confrontingambiguity6dobject, song2021scorebasedgenerativemodelingstochastic, song2020generativemodelingestimatinggradients, yang2024diffusionmodelscomprehensivesurvey} constitute a class
of probabilistic models that generate data through the reversal of a noising process. Researchers have applied diffusion models~\cite{hsiao2024confrontingambiguity6dobject, murphy2022implicitpdfnonparametricrepresentationprobability, wang2024objectposeestimationaggregation, xu20246ddiffkeypointdiffusionframework} to tasks such as 6D pose estimation, which requires the estimation of an object's rigid transformation. A key challenge in object pose estimation involves the management of pose ambiguity~\cite{manhardt2019explainingambiguityobjectdetection, Park_2019, thalhammer2022copeendtoendtrainableconstant, hsiao2024confrontingambiguity6dobject}, as symmetric objects can present multiple valid solutions. Several recent works~\cite{hsiao2024confrontingambiguity6dobject, murphy2022implicitpdfnonparametricrepresentationprobability, DBLP:conf/wacv/HoferKMZ23} have introduced probabilistic models specifically designed to address this challenge. These approaches leverage a score-based diffusion method on $SE(3)$ and employ a variant of the Geodesic Random Walk \cite{DBLP:conf/nips/BortoliMHTTD22} to mitigate symmetry-induced ambiguities and enhance the robustness of pose predictions. Nevertheless, these methods face performance limitations due to their sequential nature, which prohibits parallelization and thus often results in extended computational latency.

To address the non-parallelizable denoising process, methods like DDIM~\cite{song2022denoisingdiffusionimplicitmodels} and DPMSolver \cite{lu2022dpmsolverfastodesolver} reduce the number of denoising steps via coarser discretizations, trading sample quality for speed. In contrast, ParaDiGMS~\cite{shih2023parallelsamplingdiffusionmodels} investigates whether a method exists to balance computation and speed without sample quality degradation. This approach applies the Picard iteration~\cite{berinde2007iterative}, a fixed-point iteration for solving ODEs, to restructure the sampling process in diffusion models. Rather than adhere to a strictly sequential computation graph, ParaDiGMS introduces a parallelizable structure that enables more efficient sampling. Unlike standard parallelization, which improves throughput through simultaneous generation of multiple samples, ParaDiGMS aims to reduce sample latency. This reduction minimizes the time required to produce a single sample. However, the method proposed in ParaDiGMS accelerates image generation~\cite{ho2020denoisingdiffusionprobabilisticmodels, song2022denoisingdiffusionimplicitmodels, lu2022dpmsolverfastodesolver}, where data resides in Euclidean space. In contrast, LiePoseDiffusion~\cite{hsiao2024confrontingambiguity6dobject} addresses the pose ambiguity problem on the $SO(3)$ manifold. To the best of our knowledge, no existing work explores Picard iteration on the ${SO}(3)$ manifold to address pose estimation and acceleration.

Motivated by the parallelizable characteristics of ParaDiGMS, this study aims to accelerate the diffusion process for 6D pose estimation through numerical Picard iteration adapted to the $SO(3)$ manifold. To this end, we provide a mathematical derivation that interprets the forward diffusion process on $SO(3)$ as a stochastic differential equation (SDE) in the Lie algebra, and demonstrate the efficient solution of the corresponding ordinary differential equation (ODE) via Picard iteration. This approach enables parallel computation across time steps, which significantly reduces sampling latency. We evaluate our method on the SYMSOL dataset~\cite{murphy2022implicitpdfnonparametricrepresentationprobability}, which consists of images of symmetric objects. The experimental evidence demonstrates that our algorithm achieves up to a $4.9\times$ speed-up, which significantly reduces the latency of generating a single sample. Without any model retraining, our method achieves the same minimum angular distance as 
LiePoseDiffusion~\cite{hsiao2024confrontingambiguity6dobject}. Moreover, we provide qualitative results that illustrate the accurate capture of symmetric pose distribution of objects by our method. Our main contributions can be summarized as follows:
\vspace{-1.5em}
\begin{itemize}[itemsep=0pt, parsep=0pt]
    \item We apply a numerical method on the ${SO}(3)$ manifold to solve the probability flow ODE and propose a parallelizable algorithm. Our method reliably captures both continuous and discrete symmetries, as demonstrated by the visualized sample results.
    \item By exploiting parallelism in the Picard iteration, our algorithm achieves up to a $4.9\times$ speedup over the sequential sampling method, that reduces latency without necessitating any model retraining.
\end{itemize}


\begin{figure*}[t]
    \vspace{-3.5em}
    \centering
    \includegraphics[width=\textwidth]{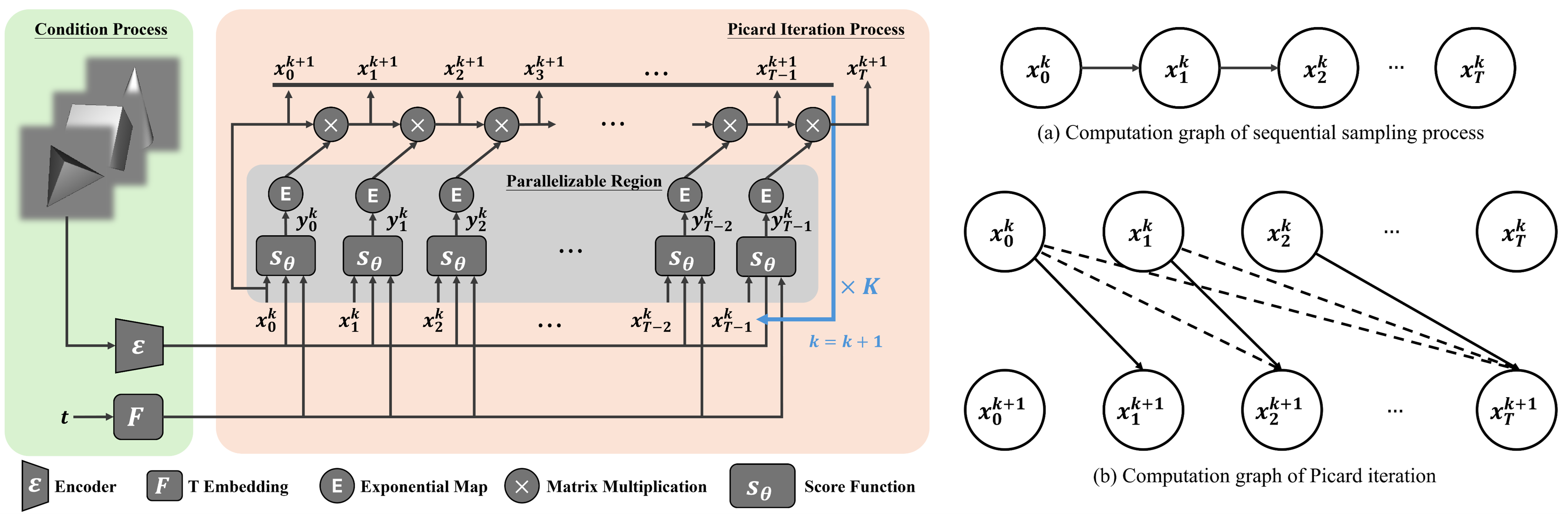}
    \vspace{-2em}
    \caption{
    \textbf{Left:} Overview of the proposed framework. The diffusion model conditions on the input image and the timestep $t$; the image is first processed by an encoder, while $t$ is embedded using a sinusoidal function. 
    \textbf{Right:} (a) Computation graph of the sequential geodesic random walk, where each update depends on the previous timestep. (b) Parallelizable computation graph based on Picard iteration, where the update at iteration \(k+1\) depends only on values from iteration \(k\), and is independent across different timesteps.}
    \label{fig:architecture and computation_graph}
    \vspace{-2.0em}
\end{figure*}

\vspace{-1.0em}
\section{Preliminary}
\vspace{-0.7em}
\subsection{Lie Group}
\vspace{-0.7em}
A Lie group~\cite{sola2018micro}, denoted as \( \mathcal{G} \), is a mathematical structure that combines the properties of both a group and a smooth manifold. A smooth manifold is a space that locally resembles Euclidean space, enabling calculus. As a group, \( \mathcal{G} \) satisfies the group axioms, with an operation defined by a map \( \circ: \mathcal{G} \times \mathcal{G} \to \mathcal{G} \) that combines two elements. The product of two elements \( X, Y \in \mathcal{G} \) is typically denoted as \( X \circ Y \), which we will write \( XY\) for convenience. Each Lie group \( \mathcal{G} \) has an associated Lie algebra \( \mathfrak{g} \), connected via the exponential map \( \text{Exp}: \mathfrak{g} \to \mathcal{G} \) and the logarithmic map \( \text{Log}: \mathcal{G} \to \mathfrak{g} \). In the 6D pose estimation domain, \( \textit{SO}(3) \) and \( SE(3) \) are the most commonly used Lie groups. 

\vspace{-0.7em}
\subsection{Stochastic Differential Equation (SDE)}
\vspace{-0.7em}
%
Prior endeavors~\cite{ANDERSON1982313, song2021scorebasedgenerativemodelingstochastic} show that both the forward and reverse diffusion processes can be formulated as SDEs. The reverse-time SDE involves the score function $\nabla_\mathbf{x} \log p_t(\mathbf{x})$, which directs the trajectory toward high-density regions of the data distribution. These processes are governed by the following equations:
\vspace{-0.7em}
\begin{equation}
    d\mathbf{x} = f(\mathbf{x}, t)\,dt + g(t)\,d\mathbf{w},
\label{eq:forward-SDE}
\end{equation}
\vspace{-1.8em}
\begin{equation}
    d\mathbf{x} = \left(f(\mathbf{x}, t) - g(t)^2 \nabla_\mathbf{x} \log p_t(\mathbf{x})\right) dt + g(t) d\bar{\mathbf{w}},
\label{eq:reverse-SDE}
\vspace{-0.1em}
\end{equation}
where $\mathbf{w}$ represents the standard Wiener process in forward time, and $\bar{\mathbf{w}}$ denotes its reverse-time counterpart. Here, $f(\mathbf{x}, t)$ is the drift coefficient, and $g(t)$ is the diffusion coefficient. For any diffusion process, there exists a corresponding deterministic process whose trajectories share identical marginal probability densities $\{p_t(\mathbf{x})\}_{t=0}^T$ with the SDE in Eq.~(\ref{eq:reverse-SDE}). This process satisfies the probability flow ODE, given by:
\vspace{-0.6em}
\begin{equation}
\resizebox{0.43\textwidth}{!}{
$d\mathbf{x} = \left(f(\mathbf{x}, t) - \frac{1}{2} g(t)^2 \nabla_\mathbf{x} \log p_t(\mathbf{x})\right) dt,\quad\mathbf{x}_t \sim \mathcal{N}(\mathbf{0}, \mathbf{I}).
$}
\label{eq:reverse-ODE}
\vspace{-0.6em}
\end{equation}
Generalizing score-based diffusion models to an SDE and converting them into a probability flow ODE constitutes a critical aspect in previous works \cite{song2022denoisingdiffusionimplicitmodels, lu2022dpmsolverfastodesolver, song2021scorebasedgenerativemodelingstochastic}, which enables efficient sampling with reduced denoising steps. This transformation is equal importance in our work, allowing the use of numerical ODE solvers such as Euler--Maruyama methodology or Picard iteration.

\vspace{-0.7em}
\subsection{Picard Iteration}
\vspace{-0.7em}
Picard iteration~\cite{berinde2007iterative} serves as a common method to solve ODEs via fixed-point iteration. ParaDiGMS~\cite{shih2023parallelsamplingdiffusionmodels} proposed this approach to accelerate sampling in diffusion models. Consider the ODE formulated as follows:
\vspace{-1.0em}
\begin{equation}
    d\mathbf{x}(t)=s(\mathbf{x}(t),t)dt,
\label{eq:simple-ODE form}
\vspace{-0.3em}
\end{equation}
where \( s(\mathbf{x}(t), t) \) denotes the drift coefficient.

This equation can be iteratively solved through initialization of a guess $\{\mathbf{x}^k(t):0\le t\le 1\}$ at $k=0$. The update formula takes the form expressed as follows:
\vspace{-0.7em}
\begin{equation}
    \mathbf{x}^{k+1}(t)=\mathbf{x}^k(0)+\int^t_0s(\mathbf{x}^k(u),u)du.
\label{eq:picard iteration}
\vspace{-0.7em}
\end{equation}
The Picard-Lindelöf theorem~\cite{berinde2007iterative} ensures that these iterates form a convergent sequence, while the Banach fixed-point theorem~\cite{berinde2007iterative} guarantees convergence to the unique ODE solution with initial value $\mathbf{x}(0)$. We discretize Eq.~(\ref{eq:picard iteration}) with a step size $\frac{1}{T}$ for $t\in [0,T]$:

\vspace{-1.0em}
\begin{equation}
    \mathbf{x}^{k+1}_t=\mathbf{x}^k_0+\frac{1}{T}\sum_{i=0}^{t-1}s(\mathbf{x}^k_i,i/T).
\label{eq:discrete picard}
\vspace{-0.8em}
\end{equation}
From Eq.~(\ref{eq:discrete picard}), we identified the two key observations.
\begin{enumerate}[itemsep=0.1em, topsep=0.2em]
    \item The computation for each timestep operates independently, which enables potential parallelization.
    \item Even in worst-case scenarios, the exact convergence occurs in $K \le T$ iterations, where $ K $ is the number of iterations required for convergence. In practice, the required iterations for convergence falls substantially below $T$, as proven in \cite{shih2023parallelsamplingdiffusionmodels}.
\end{enumerate} 
To illustrate the first property, we visualized the computation graph of Eq.~(\ref{eq:discrete picard}). Fig.~\ref{fig:architecture and computation_graph} (right, subfigure (b)) demonstrates that the $(k+1)$-th update depends exclusively on results from iteration $k$. This structure permits parallel computation across different timesteps. In contrast, the original geodesic random walk, shown in Fig.~\ref{fig:architecture and computation_graph}~(right, subfigure (a)), requires sequential computation due to its timestep-dependent update structure.

\vspace{-0.7em}
\section{Methodology}
\vspace{-0.7em}
In this section, we demonstrate the application of Picard iteration to the Lie group ${SO}(3)$ and derive a parallelizable update equation for sampling elements from diffusion models. We first establish that the perturbation kernel on ${SO}(3)$ can be interpreted as an SDE in the Lie algebra. Subsequently, by transforming the problem to the Lie algebra, which can be treated as a Euclidean space, we apply Picard iteration to solve the corresponding ODE.
To begin with, the perturbation kernel employed in LiePoseDiffusion on the ${SO}(3)$ manifold can be formally expressed as the following:
\vspace{-0.5em}
\begin{equation}
\resizebox{0.4\textwidth}{!}{$
    p_\Sigma(Y|X)=\frac{1}{\zeta(\Sigma)} \exp \left(-\frac{1}{2}\text{Log}(X^{-1}Y)^T\Sigma^{-1}\text{Log}(X^{-1}Y)\right), 
$}
\label{eq::eq7}
\vspace{-0.5em}
\end{equation}
where \( \zeta(\cdot) \) denotes the normalizing constant, and $ X, Y $ are rotation matrices on the $SO(3)$ manifold. Assume that \(\Sigma=\sigma^2 \mathbf{I}\), where \(\sigma\) is the standard deviation and \( \mathbf{I} \) denotes the identity matrix, and let \(Y=X_i, X=X_0\), Eq.~(\ref{eq::eq7}) can be re-written as the following formulation:
\vspace{-0.4em}
\begin{equation}
\resizebox{0.4\textwidth}{!}{$
p_{\sigma^2 I}(X_i \mid X_0) = \frac{1}{\zeta(\sigma^2 I)}\exp\left( -\frac{1}{2\sigma^2} 
\left\lVert \text{Log}(X_i^{-1}X_0) \right\rVert_2^2 \right)
.$}
\label{eq:perturbation kernel diagonal covariance}
\vspace{-0.4em}
\end{equation}
The $\exp(\cdot)$ in Eq.~(\ref{eq:perturbation kernel diagonal covariance}) refers to the standard exponential function, not the exponential mapping $\text{Exp}(\cdot)$ used for mapping between a Lie algebra and its corresponding Lie group. Eq.~(\ref{eq:perturbation kernel diagonal covariance}) represents a Gaussian distribution with diagonal covariance. The logarithmic transformation $z = \text{Log}(X_0^{-1} X_i)$ characterizes the relative rotation between elements in the manifold, and it follows the gaussian distribution \( z \sim \mathcal{N}(\mathbf{0}, \sigma^2 \mathbf{I}) \). Under the assumption that $X_0$ approximates $X_i$ such that $X_0^{-1}X_i \approx \mathbf{I}$, this logarithmic term permits approximation as $\text{Log}(X_0^{-1} X_i) \approx \text{Log}(X_i) - \text{Log}(X_0)$. This approximation leads to the stochastic formulation:
\vspace{-0.7em}
\begin{equation}
    \text{Log}(X_i)-\text{Log}(X_0)=\sigma\epsilon,
    \label{eq:SMLD}
\vspace{-0.5em}
\end{equation}
%
where \(\epsilon=\mathcal{N}(\mathbf{0},\mathbf{I}) \). Let $y_i=\text{Log}(X_i)$, and $y_i=y_0+\sigma_i \epsilon$, which defines the SMLD perturbation kernel~\cite{song2020generativemodelingestimatinggradients} in the Euclidean space. Research by~\cite{song2021scorebasedgenerativemodelingstochastic} demonstrates that this formulation converges to the Variance Exploding SDE (VE-SDE) as:
\vspace{-1.2em}
\begin{equation}
    dy=\sqrt{\frac{\sigma^2(t)}{dt}}dw.
    \label{eq:SMLD}
\vspace{-0.6em}
\end{equation}
The derivation above establishes that the diffusion process on ${SO}(3)$ can be interpreted as an SDE in the Lie algebra, which exhibits Euclidean space properties. Eq.~(\ref{eq:SMLD}) takes the form of Eq.~(\ref{eq:forward-SDE}), where $f(t)=0$ and $g(t)=\sqrt{\frac{\sigma^2(t)}{dt}}$. The corresponding reverse-time SDE emerges from Eq.~(\ref{eq:reverse-SDE}), which then transforms into a probability flow ODE via Eq.~(\ref{eq:reverse-ODE}). This ODE yields to solution through Picard iteration. Finally, Eq.~(\ref{eq:discrete picard}) produces the discretized update rule as the following:
\vspace{-1em}
\begin{equation}
    y^{k+1}_t=y^k_0+\frac{1}{T}\sum_{i=0}^{t-1}-\frac{1}{2}g^2(i/T)\nabla_y \log p_i(y_i).
    \label{eq:proof-discretize update}
\vspace{-0.8em}
\end{equation}
The relation $y_t=\text{Log}(X_t)$, when substituted into Eq.~(\ref{eq:proof-discretize update}) and rearranged followed by the exponential mapping to both sides, produces the update equation:
%
%
\begin{algorithm}[h]
\scriptsize 
\caption{\footnotesize{Parallel Sampling via Picard Iteration on  \(\textit{SO}(3)\)}}
\label{alg:picard_so3}

\begin{algorithmic}[1]
    \Statex \textbf{Input:} Diffusion model \(s_\theta(\cdot, \cdot)\) with variances \(\sigma_t\), tolerance \(\tau\), step count \(T\), batch window size \(p\), and coefficient \(c_t\).
    \Statex \textbf{Output:} A sample from \(s_\theta\).
    \State t,k\(\leftarrow\)0,0
    \State  \(\mathbf{X}^k_0\)=\text{Exp}\((\epsilon)\),\quad \(\epsilon \sim \mathcal{N}(\mathbf{0}, \mathbf{I}),\quad \mathbf{X}^k_i\leftarrow\mathbf{X}^k_0 \quad \forall i\in [1,p]\).
    \While{\(t < T\)}
        \State \(y_{t+j}\leftarrow s_\theta(\mathbf{X}_i^k,\sigma_t)\quad \forall i\in [0,p)\)
        \State \(\mathbf{X}^{k+1}_{t+j+1}\leftarrow  \mathbf{X}^k_t \prod_{i=t}^{t+j}c_i \text{Exp}(\mathbf{y}_i) \quad \forall j \in [0,p)\)
        \State error \(\leftarrow \{ \lVert \mathbf{X}^{k+1}_{t+j}-\mathbf{X}^k_{t+j} \rVert:\forall j\in [1,p)\}\)
        \State stride \(\leftarrow\min(\{j:\text{error}_j>\tau^2\sigma_j^2\}\cup \{p\})\)
        \State \(\mathbf{X}_{t+p+j}^{k+1}\leftarrow \mathbf{X}^{k+1}_{t+p} \quad \forall j\in [1,\text{stride}]\)
        \State \(t\leftarrow t + \text{stride}, \quad k\leftarrow k+1\)
        \State \(p\leftarrow \min(p,T-t)\)
    \EndWhile
    \State \Return \(\mathbf{x}^k_T\)
\end{algorithmic}
\end{algorithm}

\vspace{-2em}
\begin{equation}
\resizebox{0.4\textwidth}{!}{$
    X_t^{k+1} = X_0^k \prod_{i=0}^{t-1} \text{Exp} \left( \frac{-g^2(i/T)}{2T} \nabla_{\text{Log} X} \log p_i (\text{Log} X_i^k) \right).
$}
\label{eq:Final Update Equation}
\vspace{-0.3em}
\end{equation}

The final update equation in  Eq.~(\ref{eq:Final Update Equation}) enables adaptation of the algorithm from~\cite{shih2023parallelsamplingdiffusionmodels} to operate on the ${SO}(3)$ manifold. Algorithm~\ref{alg:picard_so3} presents the pseudocode, while Fig.~\ref{fig:architecture and computation_graph}~(left) illustrates the overall framework. The algorithm employs score function parameterization with $\theta$. Line~5 in Algorithm~\ref{alg:picard_so3} constitutes the most significant step, which implements the update in Eq.~(\ref{eq:Final Update Equation}). The forward pass of the neural network represented in Line~4 permits parallel computation.

\section{Experimental Results}
\vspace{-0.2em}

\begin{figure*}[t]
    \vspace{-3.5em}
    \centering
    \includegraphics[width=\textwidth]{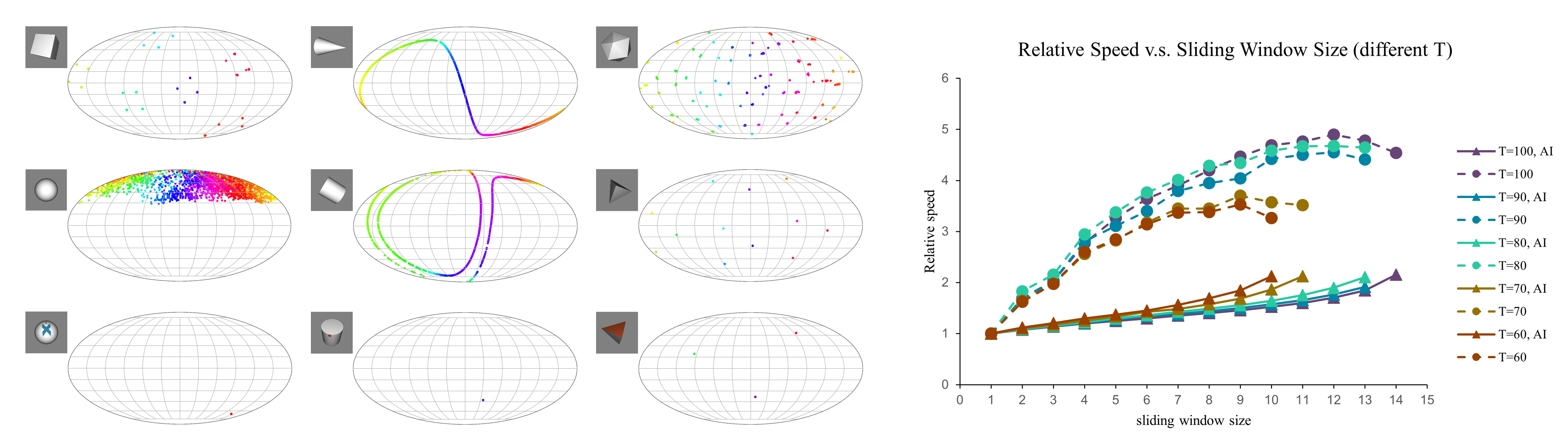}
    \vspace{-2em}
    \caption{\textbf{Left:} Visualization of 2,000 samples generated by our algorithm using images from the SYMSOL dataset. \textbf{Right:} The x-axis represents the batch window size, which determines the number of data points considered at each step. The y-axis shows the relative speed, indicating the performance improvement as a ratio compared to the baseline.}
    \label{fig:visualization and batch_window}
    \vspace{-1em}
\end{figure*}

\vspace{-0.7em}
\subsection{Experimental Setups}
\vspace{-1em}
\noindent\textbf{Dataset.}
We employ the SYMSOL dataset in accordance with \cite{murphy2022implicitpdfnonparametricrepresentationprobability, hsiao2024confrontingambiguity6dobject}. SYMSOL comprises images of symmetric objects
which serve to evaluate an algorithm's capabilities in addressing the pose ambiguity problem. SYMSOL further incorporates textured variants 
which increase task complexity by requiring texture information consideration during pose distribution prediction.

\vspace{0.5em}\noindent
\textbf{Model Architecture.} \
Following the JAX implementation of LiePoseDiffusion \cite{James2018Jax}, we re-implemented the method in PyTorch. The model accepts an RGB image $I$ and a timestep $t$ as conditional inputs. The denoising component predicts the score for the noised input $\tilde{x}_t$. The image encoder $\mathcal{E}$, constructed upon a ResNet-34 backbone \cite{DBLP:conf/cvpr/HeZRS16}, processes images with dimensions $224 \times 224$ pixels. The denoising network subsequently processes $\tilde{x}_t$, $t$, and $\mathcal{E}(I)$ to predict the final score, denoted as $s_\theta(\tilde{x}_t, t, \mathcal{E}(I))$.  
The image encoding $\mathcal{E}(I)$ remains consistent across all timesteps, thus requires computation only once for each input image.

\vspace{0.5em}\noindent
\textbf{Evaluation Method.} \
In our experiments, we evaluate our proposed algorithm through dual perspectives: performance and efficiency. For performance assessment, we report the average angular distances in degrees. For efficiency evaluation, we measure the inference time required to sample either individual or multiple objects. Our experiments operate on an Intel i7-12700H CPU and an RTX 4070 GPU. We compare our parallelizable algorithm with LiePoseDiffusion~\cite{hsiao2024confrontingambiguity6dobject}.


\vspace{-0.7em}
\subsection{Quantitative Results}
\vspace{-0.7em}
\noindent
\textbf{Time Distribution in Forward Pass.} \
Table~\ref{tab:time_distribution and accuracy_time}~(left) reports the processing time of the sequential sampling method~\cite{hsiao2024confrontingambiguity6dobject} across different components in the forward pass. The image undergoes initial processing by the encoder, followed by 100 iterations of the MLP-head denoising process. The results suggest that the MLP-head dominates computation, constituting $98.6\%$ of the total time, while the backbone contributes merely $1.4\%$. The sequential nature of MLP-head iterations reveals a substantial parallelizable region, which presents a significant opportunity for acceleration.

\vspace{0.5em}\noindent
\textbf{Total Inference Time Comparison.} \
We next present a speed comparison between the sequential algorithm and our parallelized version in Table~\ref{tab:time_distribution and accuracy_time}~(right). At 100 and 50 steps, our proposed method achieves speedups of 4.9$\times$ and 2.9$\times$, respectively, without significant degradation in minimum angular distance.

\vspace{0.5em}\noindent
\textbf{Sampling Multiple Instances.} \
We evaluate the performance for multi-sample inference. Table~\ref{tab:multiple_instances} shows that our method achieves real-time performance ($>30$ fps) with small sample sizes, which executes $7\times$ faster than the sequential version. Even with $2,000$ samples, ours maintains a $1.3\times$ speed advantage. This reduction derives from a relative increase in the number of model evaluations as compared to the sequential approach.


\vspace{-0.7em}
\subsection{Qualitative Results}
\vspace{-0.7em}
Fig.~\ref{fig:visualization and batch_window}~(left) presents the visualization results of $2,000$ samples for nine objects. The cylinder and cone exhibit continuous symmetry, whereas the tetrahedron, cube, and icosahedron possess $12$, $24$, and $60$ discrete symmetries, respectively. The remaining three objects, which constitute textured versions of the tetrahedron, cylinder, and ball, feature surface patterns that reduce symmetry when the textured side appears. For instance, the pose distribution with the object in the middle-left subfigure does not display the pattern on the ball, which results in a widely dispersed pose distribution. In contrast, the bottom-left subfigure reveals the textured pattern, which diminishes the symmetry and produces a concentrated pose distribution. Our proposed algorithm generates results that closely align with those reported in~\cite{hsiao2024confrontingambiguity6dobject}, which demonstrates the strong reliability and capability of our method.

\begin{table}[t]
\centering
\captionsetup{font=scriptsize}
\vspace{-0.8em}
\caption{\textbf{Left:} Processing time for each component in the forward pass. \textbf{Right:} Comparison of wall-clock time between the original sequential and our parallelizable algorithm across different denoising steps. \textbf{D}: Minimum angular distance. \textbf{T}: Inference time per sample.}
\vspace{-0.5em}
\begin{minipage}[t]{0.49\linewidth}
\centering
\resizebox{\linewidth}{!}{
\begin{tabular}{c c c}
\toprule
\textbf{Components} & \textbf{Time (ms)} & \textbf{Percentage (\%)} \\
\midrule
Backbone & 1.98 & 1.42 \\
Head & 138.60 & 98.59 \\
Total & 140.58 & 100.00 \\
\bottomrule
\end{tabular}
}
\end{minipage}
\hfill
\begin{minipage}[t]{0.49\linewidth}
\centering
\resizebox{\linewidth}{!}{
\begin{tabular}{c|c c|c c}
\toprule
\multirow{2}{*}{\textbf{Steps}} & \multicolumn{2}{c|}{\textbf{Sequential}} & \multicolumn{2}{c}{\textbf{Parallel (Ours)}}  \\
& \textbf{D (deg)} & \textbf{T (ms)} & \textbf{D (deg)} & \textbf{T (ms)} \\
\midrule
100 & 1.48 & 142.0 & 1.40 & \textbf{29.0} \\
50 & 1.42 & 72.6 & 1.39 & \textbf{25.3} \\
\bottomrule
\end{tabular}
}
\end{minipage}
\label{tab:time_distribution and accuracy_time}
\vspace{-2.0em}
\end{table}
\begin{table}[t]
\captionsetup{font=scriptsize}
\vspace{-0.8em}
\caption{
    \textbf{Speedup:}
    the run-time ratio between the sequential and our parallelized implementation. The best and second-best results are highlighted in \textcolor{red}{\underline{red}}, and \textcolor{blue}{blue}, respectively.
}
\vspace{-0.8em}
\centering
\resizebox{0.9\linewidth}{!}{
    \begin{tabular}{@{}l|ccc@{}}
        \toprule[1pt]
        \textbf{\# of Samples} & $\textbf{Sequential}_{/ms}$ & $\textbf{Parallel (Ours)}_{/ms}$ & \textbf{Speedup} \\
        \midrule
        10     & \textcolor{blue}{138.0} & \textcolor{red}{\underline{17.3}}  & \textbf{8.0}$\times$ \\
        50     & \textcolor{blue}{140.0} & \textcolor{red}{\underline{20.5}}  & \textbf{6.8}$\times$ \\
        100    & \textcolor{blue}{139.0} & \textcolor{red}{\underline{20.6}}  & \textbf{6.8}$\times$ \\
        2,000  & \textcolor{blue}{173.0} & \textcolor{red}{\underline{131.7}} & \textbf{1.3}$\times$ \\
        \bottomrule[1pt]
    \end{tabular}
}
\vspace{-2em}
\label{tab:multiple_instances}
\end{table}

\vspace{-0.7em}
\subsection{Ablation Studies}
\vspace{-0.7em}
We additionally conduct ablation analysis on the relationship between speedup and batch window size under different step sizes $T$. Specifically, we adopt the \textit{Algorithm inefficiency} (AI) metric from~\cite{shih2023parallelsamplingdiffusionmodels}, which quantifies the relative number of model evaluations required by Picard iteration compared to its sequential counterpart. This metric reflects the computational overhead introduced by Picard Iteration. Fig.~\ref{fig:visualization and batch_window} (right) indicates that when $T=100$, the maximum relative speed occurs with a batch window size of $12$, which yields performance approximately five times faster than the sequential version. The experiment also reveals that the AI index increases with larger batch window sizes.

\vspace{-0.7em}
\section{Conclusion}
\vspace{-0.7em}
In this paper, we proposed a mathematical formulation that interpreted the diffusion process on $\textit{SO}(3)$ as a SDE in the Lie algebra. We subsequently solved the corresponding probability flow ODE through Picard iteration and derived a discretized update equation. Based on these formulations, we developed a parallelizable algorithm that eliminated the need to retrain existing deep learning models and accelerated the denoising process through parallel computation. The experimental results demonstrated the effectiveness and efficiency of our proposed method. Compared to the baseline approach, our methodology achieved a speed-up of up to 4.9$\times$ without any degradation in accuracy.

\vspace{-0.7em}
\section{Acknowledgements}
\vspace{-0.7em}
The authors gratefully acknowledge the support from the National Science and Technology Council (NSTC) in Taiwan under grant numbers MOST 111-2223-E-002-011-MY3, NSTC 113-2221-E-002-212-MY3, and NSTC 113-2640-E-002-003. The authors would like to express their appreciation for the donation of the GPUs from NVIDIA Corporation and NVIDIA AI Technology Center (NVAITC) used in this work. Furthermore, the authors extend their gratitude to the National Center for High-Performance Computing (NCHC) for providing the necessary computational and storage resources.

\bibliographystyle{ieeetr}
\bibliography{reference}

\end{document}